%% file: main.tex
\documentclass{article}
\usepackage{iclr2025_conference, times}

\input{preamble}

\title{
    \centering
    \begin{minipage}{\textwidth} % Create a minipage to fit content in one line
        \centering
        \begin{tabular}{c c} % Create a 2-column table (one for image, one for text)
            \includegraphics[width=0.13\textwidth]{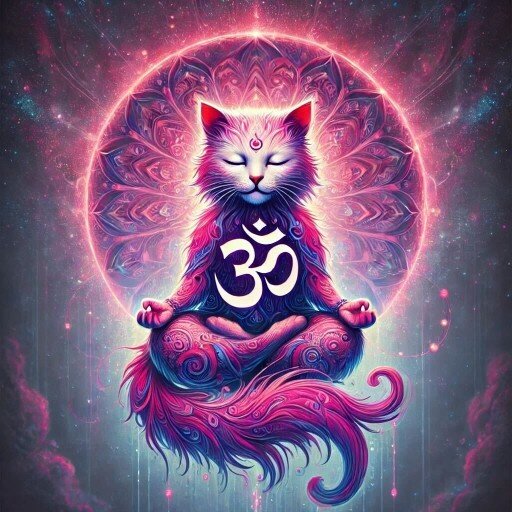} &  % Adjust width of image
            \raisebox{2.\height}{\textbf{\Large OMCAT: Omni Context Aware Transformer}} % Title text, adjusted vertically
        \end{tabular}
    \end{minipage}
}

\author{Arushi Goel$^{\dagger}$, Karan Sapra, Matthieu Le, Rafael Valle$^{*}$, Andrew Tao and Bryan Catanzaro \\    
    NVIDIA \\
    $^{\dagger}$\texttt{arushig@nvidia.com, $^{*}$rafaelvalle@nvidia.com}
}

\iclrfinalcopy % Uncomment for camera-ready version, but NOT for submission.

\begin{document}
\maketitle
\begin{abstract}
\input{abstract}
\end{abstract}

\section{Introduction} \label{sec:introduction}

\input{intro_v2}
\section{Related Work} \label{sec:related_work}
\input{related_work_v2}
\vspace{-1em}

\section{The \texttt{OCTAV} Dataset}\label{sec:dataset}
\input{dataset}

\section{The \modelname{} Approach}\label{sec:approach}
\input{approach}
\vspace{-1em}
\section{Experiments}\label{sec:experiments}
\input{experiments}
\vspace{-1em}

\section{Conclusion}\label{sec:discussion}
\input{discussion}

\clearpage
\newpage
\bibliography{references}\label{sec:references}
\bibliographystyle{iclr2025_conference}
\FloatBarrier
\clearpage
\newpage
\appendix
\section*{Appendix}
\label{sec:appendix}
\input{appendix}

\end{document}

%% file: preamble.tex
\newcommand{\modelname}{\texttt{OMCAT}\xspace}
\newcommand{\rote}{\texttt{RoTE\xspace}}

\newcommand{\eg}{\textit{e.g.}}
\newcommand{\ie}{\textit{i.e.}}
\usepackage{amsmath,amsfonts}
\usepackage{microtype}
\usepackage{graphicx}
\usepackage{subcaption}
\usepackage{booktabs} % for professional tables
\usepackage{colortbl}

\usepackage{multirow}
\usepackage{hyperref}
\usepackage{url}
\usepackage{bbding}
\usepackage{pifont}
\usepackage{wasysym}
\usepackage{amssymb}
\usepackage[capitalize,noabbrev]{cleveref}

\hypersetup{
    pdfborderstyle={/S/U/W 1}, % underline links instead of boxes
    linkbordercolor=red,       % color of internal links
    citebordercolor=green,     % color of links to bibliography
    filebordercolor=magenta,   % color of file links
   urlbordercolor=cyan        % color of external links
}

\usepackage{interval}
\usepackage{tabularx}
\usepackage{xspace}
\usepackage{mathtools}  % for conditions in math
\usepackage{siunitx}    % for units

\usepackage{placeins}
\makeatletter
\renewcommand{\paragraph}{%
  \@startsection{paragraph}{4}%
  {\z@}{0.5em}{-5em}%
  {\normalfont\normalsize\bfseries}%
}

\makeatother
\usepackage{algpseudocode}

\usepackage{algorithm}
\usepackage{amsmath}

%% file: abstract.tex
Large Language Models (LLMs) have made significant strides in text generation and comprehension, with recent advancements extending into multimodal LLMs that integrate visual and audio inputs. However, these models continue to struggle with fine-grained, cross-modal temporal understanding, particularly when correlating events across audio and video streams. We address these challenges with two key contributions: a new dataset and model, called \texttt{OCTAV} and \modelname respectively. \texttt{OCTAV} (\textbf{O}mni \textbf{C}ontext and \textbf{T}emporal \textbf{A}udio \textbf{V}ideo) is a novel dataset designed to capture event transitions across audio and video. Second, \modelname (\textbf{O}mni \textbf{C}ontext \textbf{A}ware \textbf{T}ransformer) is a powerful model that leverages \rote~(Rotary Time Embeddings), an innovative extension of RoPE, to enhance temporal grounding and computational efficiency in time-anchored tasks. Through a robust three-stage training pipeline—feature alignment, instruction tuning, and \texttt{OCTAV}-specific training—\texttt{OMCAT} excels in cross-modal temporal understanding. Our model demonstrates state-of-the-art performance on Audio-Visual Question Answering (AVQA) tasks and the \texttt{OCTAV} benchmark, showcasing significant gains in temporal reasoning and cross-modal alignment, as validated through comprehensive experiments and ablation studies. Our dataset and code will be made publicly available. The link to our demo page is \url{https://om-cat.github.io}.

%% file: intro_v2.tex
\begin{figure*}[h]
    \centering
    \includegraphics[width=\linewidth]{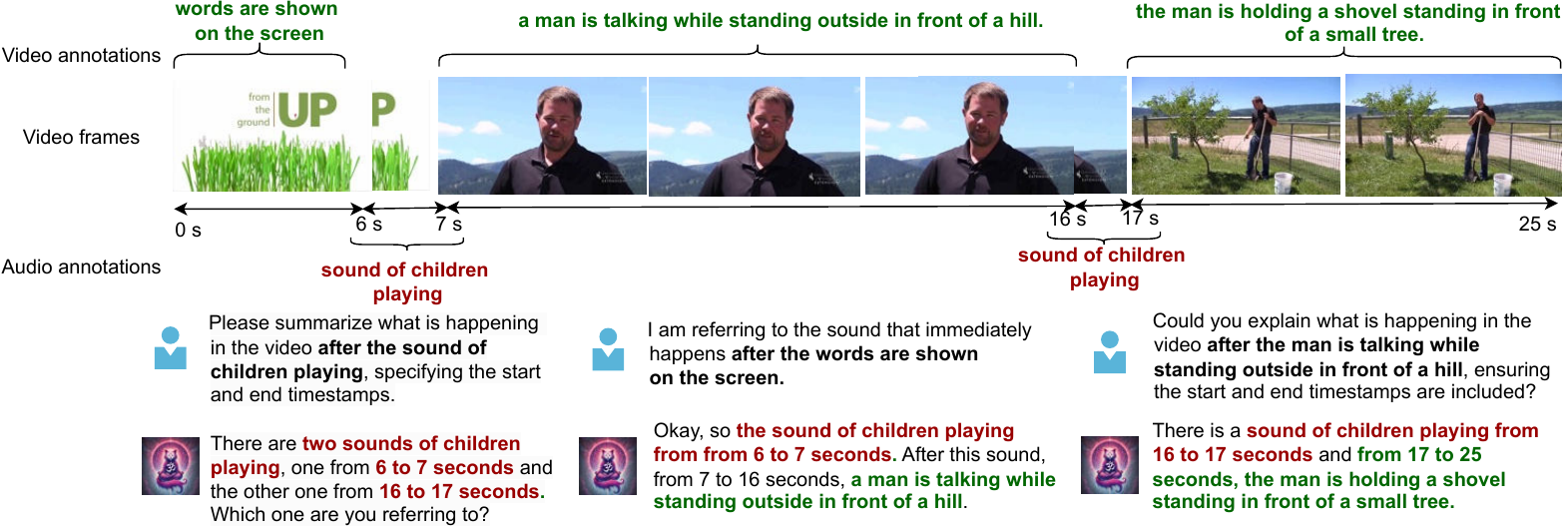}
    \vspace{-0.5em}
    \caption{Illustration of a video sequence from our proposed \texttt{OCTAV} dataset. The annotations highlight key moments, including the timing of the audio and visual events. }
    \label{fig.intro_figure}
\end{figure*}

% \arushi{Elaborate here how time is incorporated in video modality separately and audio modality separately, could also discuss more in related work}

Large language models (LLMs)~\citep{achiam2023gpt,touvron2023llama} have achieved remarkable breakthroughs in both text generation and comprehension~\citep{mckeown1992text,achiam2023gpt} tasks. Since then, significant progress has been made to extend LLMs to multimodal LLMs~\citep{cheng2024videollama,li2023videochat,maaz2023video,li2024groundinggpt}, which integrate visual and audio inputs with textual instructions to provide understanding in multimodal contexts~\citep{yang2022avqa,chen2023valor,chen2023vast}. These models, however, are limited in their cross-modal understanding and in their ability to provide answers to questions with fine-grained timestamps or anchored on events, as shown in \cref{fig.intro_figure}. In this paper, we address these limitations by proposing a new dataset \texttt{OCTAV} and a model called \modelname. The Omni Context and Temporal Audio Video dataset, \texttt{OCTAV}, consists of question-answer pairs for a video. Each question captures the transition between the events happening in the video through a sound event (\eg~\cref{fig.intro_figure}). The Omni Context Aware Transformer, \modelname, addresses the limitations of existing models~\citep{maaz2023video,tang2024avicuna,su2023pandagpt,cheng2024videollama} through a unified audio and visual language model by effectively incorporating time representations to ground the modalities temporally.
     
% Limitations of current multimodal LLMs
Despite the notable progress in multimodal LLMs~\citep{li2023videochat,maaz2023video,cheng2024videollama,lyu2023macaw}, most advancements have been centered around developing domain specific models in isolation, typically Video LLMs~\citep{wang2023internvid,fu2024video} or Audio LLMs~\citep{gong2023listen,kong2024audio,chu2023qwen}. 
However, these models still face challenges in handling fine-grained, cross-modal temporal understanding when both audio and video are provided. For instance, if a user asks the question, ``Is it raining in the video?'' This question can be answered by either just looking at the video or listening to the audio. However, as shown in \cref{fig.intro_figure}, if the user asks the question, ``Describe what happens in the video after the sound of children playing?'', the model must understand both modalities because the sound of \texttt{children playing} cannot be seen, only heard, and \texttt{what the man is doing} cannot be heard, only seen. Achieving this is challenging due to several reasons, including the lack of temporally aligned cross-modal datasets, unified models and benchmarks, and clear understanding of how to combine modalities effectively. 

% What do we exactly do ? 
Our goal is to achieve this cross-modal temporal understanding, and to this end we propose an instruction tuning dataset called \texttt{OCTAV}: \textbf{O}mni \textbf{C}ontext and \textbf{T}emporal \textbf{A}udio \textbf{V}ideo. \Cref{fig.intro_figure} shows a sample from our proposed \texttt{OCTAV} dataset. Existing audio and video understanding datasets~\citep{chen2023vast,chen2023valor,chen2020vggsound,geng2023dense} only focus on open-ended question answering tasks~\citep{yang2022avqa,li2022learning} for audio-visual events. They lack the ability to temporally ground events or describe events that involve ambiguity or missing information in one of the modalities. 
% These properties are desirable for audio and video understanding models.
Specifically, we create question-answer pairs for a video such that each question captures the transition between the events happening in the video through a sound event. For instance, as shown in \cref{fig.intro_figure}, we add the sound event of \texttt{children playing} to the silent input video between 6 to 7 seconds, during which nothing substantial happens in the video. Then, we capture the video event \texttt{before 6 seconds} and \texttt{after 7 seconds} while using the \texttt{sound of children playing} as a transition event. This setting encourages the model to not only understand the relationship between the audio and the video, but also a strong temporal understanding of both the audio and video domains in a single setup. Despite this artificial setup, our experiments show that a model trained with this data performs well in naturally occurring video and audio pairs.
% To mitigate these issues, we propose the Omni Context Temporal Audio Video dataset, \texttt{OCTAV}, and propose a method for dataset creation.

While dataset design is necessary, it is not a sufficient condition to achieve cross-modal understanding given the challenges in modelling such data. As such, we propose a new approach that embeds absolute and relative temporal information in the audio and visual features, improving the model's ability to become temporally-aware. With the goal of improving cross-modal and temporal understanding, and following common practice in multimodal LLMs~\citep{li2023videochat,cheng2024videollama,li2024groundinggpt,tang2024avicuna,fu2024video}, we divide model training into 3 stages. The first training stage is focused on feature alignment, and uses audio-text, video-text, and audio-video-text data~\citep{liu2024visual,mei2024wavcaps,chen2023vast}. In the second stage, the model is instruction-tuned with data~\citep{luo2023valley,li2023videochat,drossos2020clotho,chen2020vggsound} that promotes temporal and cross-modal understanding. Finally, the model is trained to support complex and cross-modal temporal data in the \texttt{OCTAV} dataset as shown in \cref{fig.intro_figure}. We name the model trained with our proposed \texttt{OCTAV} dataset and the temporal conditioning strategy \modelname, for \textbf{OM}ni \textbf{C}ontext \textbf{A}ware \textbf{T}ransformer. Through this learning strategy, our method outperforms existing models on AVQA tasks~\citep{yang2022avqa,li2022learning} and beats baselines by a significant margin on our proposed \texttt{OCTAV} benchmark dataset.
% Should write about datasets as well that have been designed for these tasks
 % \arushi{you can do better in the sentences above.} 
% \arushi{add about experimental settings, results and takeaways }
% Our main contributions and takeaways. 

In summary, our main contributions are as follows: \\
- We introduce a novel method for generating synthetic instruction-tuning dataset, \texttt{OCTAV}, which has temporal and contextual audio and video question/answer pairs addressing the limitations of existing datasets. This dataset has both training and evaluation samples to promote research in this direction.  \\
- We propose \modelname: a unified, temporally-aware audio and visual language model with fine-grained and cross-modal understanding, achieved through a staged training strategy that leverages all combinations of audio, video and text data. \\
- We propose \rote{}: a simple yet efficient modification to RoPE that provides better scores on benchmarks and better computational efficiency than existing approaches for temporal conditioning, especially on time-anchored tasks.\\
- Finally, we exhaustively evaluate \modelname, including ablations, on a variety of multimodal tasks. Our experiments demonstrate that our model raises the standards on AVQA tasks, temporal understanding tasks and our proposed \texttt{OCTAV} benchmark.

%% file: related_work_v2.tex
\vspace{-1em}
\noindent 
\textbf{Multimodal LLMs. }Since the rise of large language models (LLMs)~\citep{achiam2023gpt,chiang2023vicuna,touvron2023llama}, there has been growing interest in integrating additional modalities~\citep{cheng2024videollama,gong2023listen,kong2024audio}. Video LLMs~\citep{li2023videochat,fu2024video,wang2023internvid} utilize video-text datasets to address tasks like video question answering~\citep{xu2016msr,yu2019activitynet}, visual grounding~\citep{kazemzadeh2014referitgame}, and understanding temporal segments~\citep{gao2017tall,huang2024lita}. These have evolved into multimodal LLMs~\citep{cheng2024videollama,maaz2023video,lyu2023macaw}, which encode multiple modalities and focus on coarse-grained tasks like audio-video understanding and question answering~\citep{shu2023audio,chen2023valor,yang2022avqa}. However, these models struggle with fine-grained audio-visual tasks, where precise synchronization is key to deeper event comprehension.

Recent efforts have attempted to address this. GroundingGPT~\citep{li2024groundinggpt} predicts fine-grained timestamps but is limited to sound events, while AVicuna~\citep{tang2024avicuna} takes a more balanced approach to audio-visual temporal understanding. However, both models fall short in capturing intricate cross-modal temporal dynamics. Our work aims to address these gaps by focusing on fine-grained cross-modal information integration.
\vspace{-0.5em}

\noindent 
\textbf{Instruction tuning datasets.} GPT-based methods have been widely used to create datasets for video, audio, and audio-visual tasks, advancing multimodal models with large-scale resources. In video understanding, they generate and annotate datasets for tasks like video captioning~\citep{fu2024video}, video question answering~\citep{xu2016msr,yu2019activitynet}, and action recognition~\citep{yu2019activitynet}. Similarly, for audio understanding, instruction tuning datasets~\citep{kong2024audio,goel2024audio} target sound events~\citep{salamon2014dataset}, audio captioning~\citep{kim2019audiocaps}, and audio question answering~\citep{lipping2022clotho}. Recently, AI-generated datasets have also progressed in audio-visual tasks like captioning~\citep{chen2023valor}, question answering~\citep{yang2022avqa}, and dialog~\citep{alamri2019audio}.
Despite this progress, current datasets remain predominantly coarse-grained, lacking fine-grained temporal and cross-modal synchronization. Our proposed dataset, \texttt{OCTAV}, addresses this limitation, enabling more precise alignment between audio and visual cues in complex scenarios.

%% file: dataset.tex
% \arushi{rearrange the sections below}
% what is the dataset about 
% \vspace{-1em}
One of the challenges in developing models that can understand strongly timestamped and anchored events is the lack of datasets that have this information~\citep{wang2023internvid,liu2024visual,chen2020vggsound,li2023videochat,tang2024avicuna,lyu2023macaw}. To overcome this limitation, we propose a pipeline to generate a synthetic dataset called \texttt{OCTAV}, for \textbf{O}mni \textbf{C}ontext \textbf{T}emporal \textbf{A}udio \textbf{V}ideo dataset. \Cref{fig.intro_figure} shows an example from our proposed \texttt{OCTAV} dataset. First, we discuss how we identify relevant event transitions in videos. Then, we discuss how we anchor those transitions on audio samples and finally, we show how to generate question-answer pairs for these synthetically curated videos.   

% how we create this dataset / data geenration pipeline
% \vspace{-0.5em}
\noindent
\textbf{Identifying transitions between video events. }To achieve this, we utilize videos with strongly timestamped captions~\citep{zhou2018towards,krishna2017dense,tang2019coin,zala2023hierarchical}, \ie~a video $V$ with time-caption pairs $\{(t_1, c_1), (t_2, c_2) \dots (t_k, c_k)\}$, where $k$ is the total number of time chunks annotated in the video. 
Given a list of timestamped video captions indexed by $i$ and bounded by start time ($t_i^s$) and end time ($t_i^e$) each, we find pairs where the gap between end time and start time is smallest than $m$ and the sum of their lengths, from earliest to latest, is at most $T$ seconds. Empirically we set $m=10$ and $T=30$, ensuring that the videos are not too far apart and their length is not too long. Next, we discuss how to anchor sound between these video event transitions.

% Given a list of timestamps, $\{(t_1^s, t_1^e), (t_2^s, t_2^e), \cdots, 
%  (t_k^s, t_k^e)\}$, where $t_i^s$  and $t_i^e$ represent the start and end times of the i\text{-th timestamp} in the video respectively, we find pairs of timestamps $(t_i^s, t_i^e)$ and $(t_j^s, t_j^e)$ such that: 
% 1) The gap between the end time of the first timestamp and the start time of the second timestamp is less than $m$ seconds:
% $t_j^s - t_i^e < m$ and,
% 2) The combined duration of both timestamps is less than $T$ seconds.
% $(t_i^e - t_i^s) + (t_j^e - t_j^s) < T$ where, $m$ is 10 seconds such that two events $t_i$ and $t_j$ are not too far apart in the video and $T$ is 30 seconds, the maximum duration of the chunked video. Hence, the chunked videos consists of two unique events $c_i$ and $c_j$ corresponding to the timestamps $t_i$ and $t_j$ respectively. Next, we discuss how we anchor the transition between these video events on sound events. 

\vspace{-0.5em}

\noindent
\textbf{Anchoring chunked videos on a single sound event. } For these chunked videos, we inject a sound event between the timestamp $t_i^e$ and $t_j^s$. More specifically, we randomly sample a sound event $s$ from a variety of different sound sources~\citep{salamon2014dataset,fonseca2021fsd50k,piczak2015esc,rashid2023nonspeech7k}. Details of these sound sources are provided in \cref{sec.sound_events}. We remove the original audio in the given video chunk and insert this sound event between the timestamp $\{t_i^e, t_j^s\}$ to create a strongly timed video chunk anchored on a sound event. We refer to this subset of the dataset as the \texttt{OCTAV-ST} dataset where, ST is for single-turn.

\vspace{-0.5em}

\noindent
\textbf{Anchoring chunked videos on multiple sound events. } We extend the videos from a single sound event to two sound events as shown in \cref{fig.intro_figure}. Particularly, we first create a chunked video with three unique events $c_i$, $c_j$, and $c_k$ corresponding to timestamps $t_i$, $t_j$ and $t_k$ respectively, following the same procedure discussed previously. Then, we add a random sound event after removing the original audio between the timestamps $\{t_i^e, t_j^s\}$ and $\{t_j^e, t_k^s\}$. We refer to this subset with interwoven and timestamped videos with audio events as the \texttt{OCTAV-MT} dataset where, MT stands for multi-turn.

\vspace{-0.5em}

\noindent
\textbf{Creating question-answer pairs. }Here, we discuss how to create question-answer pairs for the interwoven videos in the \texttt{OCTAV-ST} and \texttt{OCTAV-MT} dataset. Essentially, we have two (or three) video caption events for each chunked video and an associated audio event/sound between the video events. The model has to generate questions such that it can capture \textit{what event is happening in the video \{before the sound event, after the sound event\}}, and \textit{clarify which of the sound events the user is referring to while answering the question}. We use GPT-assisted~\citep{achiam2023gpt} generation to generate a diverse set of question-answer pairs. The prompts used are given in \cref{sec.octav_st_prompt} and \cref{sec.octav_mt_prompt} and the list of instructions are given in the \cref{sec.instructions}.
% comparison to existing datsets 
% \vspace{-0.5em}
\begin{table*}[!h]
    \centering
    \begin{minipage}{0.46\linewidth}
        \small
        \centering
        \caption{Statistics with number of videos and question-answer pairs for the \texttt{OCTAV-ST} dataset.}
        \renewcommand*{\arraystretch}{1.0}
        \resizebox{\linewidth}{!}{
        \begin{tabular}{c|cc}
            & Train& Test \\ \toprule
            \texttt{OCTAV-ST} & \#Videos (QA Pairs) & \#Videos(QA Pairs) \\ \toprule
            Youcook2~\citep{zhou2018towards} & 6832 &  2414 \\
            ActivityNet~\citep{krishna2017dense} & 16072 & 6228  \\
            QueryD~\citep{oncescu2021queryd} & 16985 &  - \\
            COIN~\citep{tang2019coin} & 31938 &  - \\
            HiREST~\citep{zala2023hierarchical} & 2408 & -  \\
            \midrule
            Total & 127,507 & 8642  \\
             \bottomrule    
        \end{tabular}}
        \label{tab.stats1}
    \end{minipage}
    \hspace{0.04\linewidth}
    \begin{minipage}{0.48\linewidth}
        \small
        \centering
        \caption{Statistics with number of videos and question-answer pairs for the \texttt{OCTAV-MT} dataset.}
        \renewcommand*{\arraystretch}{1.0}
        \resizebox{\linewidth}{!}{
        \begin{tabular}{c|cc}
            & Train& Test \\ \toprule
            \texttt{OCTAV-MT} & \#Videos, \#QA Pairs & \# Videos, \#QA Pairs \\ \toprule
            Youcook2~\citep{zhou2018towards}  & 4296, 34330 & 1476, 11806  \\
            ActivityNet~\cite{krishna2017dense}  &6463, 51670  & 1362, 10858  \\
            UnAV-100-MT & 14698, 94916  & 2043, 9694  \\
            % AVSD~\citep{} &  7985, 159700 & 1863, 37260  \\
            \midrule
            Total &25,457, 180,916 &  4,881, 32,358 \\
             \bottomrule    
        \end{tabular}}
        \label{tab.stats2}
    \end{minipage}
\end{table*}

% statistics of the dataset
\noindent
\textbf{Dataset Statistics. }
We utilize timestamped videos from Youcook2~\citep{zhou2018towards}, QueryD~\citep{oncescu2021queryd}, ActivityNet~\citep{krishna2017dense}, COIN~\citep{tang2019coin}, UnAV-100~\citep{geng2023dense} and, HiREST~\citep{zala2023hierarchical} datasets to create chunked videos. Essentially, we use these datasets as they have segmented annotations available for videos in diverse domains such as cooking, daily activities, scenes and instructional videos. 

Overall, the \texttt{OCTAV-ST} dataset has 127,507 unique videos with single question-answer pairs for each video for training. For evaluation, we provide 2414 unique videos with question-answer pairs from the test subset of Youcook2~\citep{zhou2018towards}, denoted as \texttt{OCTAV-ST}-Youcook2 and 6228 unique videos with question-answer pairs from the test subset of the ActivityNet dataset~\citep{krishna2017dense}, called as \texttt{OCTAV-ST}-ActivityNet. In \cref{tab.stats1}, we show the breakdown of our proposed \texttt{OCTAV-ST} dataset in detail.

The \texttt{OCTAV-MT} dataset has 25,457 unique videos/multi-turn dialogues with a total of 180,916 single question-answer pairs for training. In \cref{tab.stats2}, we show the detailed statistics of our proposed \texttt{OCTAV-MT} dataset. Specifically, we curate synthetic chunked videos for Youcook2 and ActivityNet and use the original videos from UnAV-100 dataset~\citep{geng2023dense}. The UnAV-100 dataset has timestamped audio-visual annotations from videos with real-time audio events and we convert this into question-answer pairs called the \texttt{OCTAV-MT} dataset (\eg~shown in \cref{fig.octav_mt_unav}). We train and evaluate on this dataset to show \modelname's performance on in-the-wild and naturally occurring audio-visual settings. For evaluation on this multi-turn setup, we provide a total of 4818 unique videos with 32,358 question-answer pairs. Example annotations from both the \texttt{OCTAV-ST} and \texttt{OCTAV-MT} are given in \cref{sec.examples_octav}.

\begin{table*}[!h]
    \small
    \centering
    \caption{Comparison of our proposed \texttt{OCTAV} dataset with other datasets with respect to modalities (audio/video), caption availability, multi-turn setup and timestamp information. }
    \renewcommand*{\arraystretch}{1.0}
    \resizebox{0.7\linewidth}{!}{
    \begin{tabular}{l|ccccc}
        \toprule
        Dataset & Audio & Video & Detailed captions & Multi-turn &Timestamps \\ \midrule
        InternVid~\citep{wang2023internvid} & \ding{55} &  \ding{51}& \ding{51} & \ding{51} & \ding{51}  \\
        VALOR~\citep{chen2023valor} & \ding{51} &  \ding{51}& \ding{51} & \ding{55} & \ding{55}  \\
        VAST~\citep{chen2023vast} & \ding{51} & \ding{51} & \ding{51} & \ding{55} &  \ding{55} \\ 
        VGG-Sound~\citep{chen2020vggsound} & \ding{51} &\ding{51}  &\ding{55} &  \ding{55}& \ding{55}  \\ 
        UnAV-100~\citep{geng2023dense} & \ding{51} & \ding{51} & \ding{55} & \ding{55} & \ding{51}  \\ \midrule
        % PU-VALOR~\citep{} & \ding{51} & \ding{51} & \ding{51} & \ding{55} &  \ding{51} \\ \midrule
        \texttt{OCTAV} & \ding{51} & \ding{51} & \ding{51} & \ding{51} &  \ding{51} \\ 
                 \bottomrule    
        \end{tabular}}
    \label{tab:dataset_comparison}
\end{table*}

% \vspace{-0.5em}
\noindent
\textbf{Comparison to existing datasets}
% \label{sec.comparison}
In \cref{tab:dataset_comparison}, we compare our proposed \texttt{OCTAV} dataset to existing datasets in the audio and video domains. Most of these datasets are limited to either the video modality~\citep{wang2023internvid}, have missing timestamp information~\citep{chen2023valor,chen2023vast,chen2020vggsound}, do not offer multi-turn question-answer pairs~\citep{chen2023valor,chen2023vast,chen2020vggsound,geng2023dense} or have single event classes rather than detailed captions~\citep{chen2020vggsound,geng2023dense}. \texttt{OCTAV} dataset addresses all the above mentioned limitations and provides a comprehensive benchmark for interwoven and fine-grained audio-visual understanding.  
% \vspace{0.5em}

%% file: approach.tex
\begin{figure*}[!h]
    \centering
    \includegraphics[width=\linewidth]{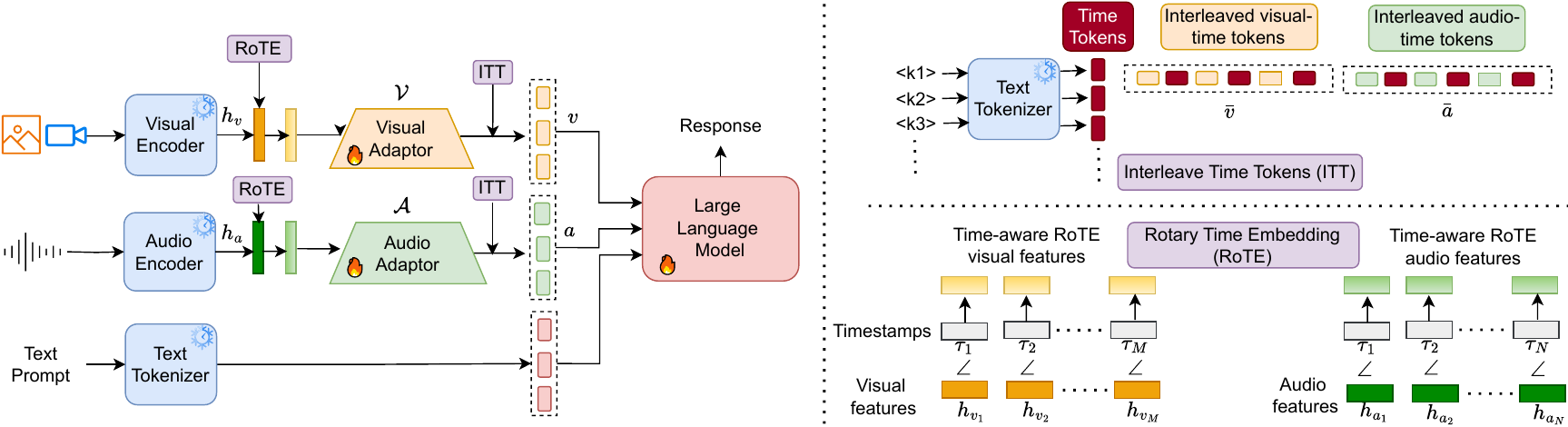}
    \caption{Overview of the \modelname pipeline. Video frames are processed through a frozen visual encoder, while audio frames are encoded using a frozen audio encoder. Extracted features are fine-tuned through adaptor layers across all three stages. The LLM remains frozen in Stage 1 and is fine-tuned in Stages 2 and 3. The purple blocks represent time alignment modules, with only one of them activated during training. $\angle$ in bottom right denotes the rotation angle.}
    \label{fig.omcat}
\end{figure*}

% \vspace{-0.5em}

In this section, we describe our proposed \modelname model, depicted in \cref{fig.omcat}. We begin by discussing the model architecture and feature extraction in \cref{sec.model}, followed by time alignment between audio and video in \cref{sec.time}.
% where we discuss extraction of features from each modality, followed by how time representations are incorporated and how audio visual adaptors are used to map the modality space to LLM embedding space. 
% Finally, we discuss the Large Language Model (LLM) backbone. 
Next, we discuss the prompt design to query the LLM in \cref{sec.prompts} and finally, we detail the multi-stage training process of \modelname in \cref{sec.multistage}.

\subsection{Model Architecture and Feature Extraction}
\label{sec.model}

% \matthieu{Do you use Rope on top of RoTE?}
\vspace{-0.5em}

\paragraph{Multi-modal Feature Extraction. }
As shown in \cref{fig.omcat}, \modelname uses a visual encoder, $f_v(.)$ and an audio encoder, $f_a(.)$. Given a video $V$ and an audio $A$, the encoded hidden features for the two modalities are represented as:
\begin{equation}
\label{eq.mmencoder}
    h_v = f_v(V), ~~~~ h_a = f_a(A)
\end{equation}
where $h_v \in \mathbb{R}^{M \times {d_v}}$ are the extracted features for the video modality with $M$ frames extracted uniformly from the video and $d_v$ as the hidden dimension. $M$ is 1 if the modality is image. The features for the audio modality are denoted as $h_a \in \mathbb{R}^{N \times {d_a}}$, where $N$ are the time windows for which the audio features are computed and $d_a$ is the hidden dimension. 

\vspace{-1em}

% \karan { we should start mappping / referring back to the pic. }
\paragraph{Audio-Visual Adaptors. }To map the video modality and audio modality to the text embedding space of the LLM~\citep{chiang2023vicuna}, we use two adaptor blocks: one for the video modality denoted as $\mathcal{V}(.)$ and another for the audio modality denoted as $\mathcal{A}(.)$. Essentially, the encoded hidden features are passed to the adaptors to extract token embeddings as:
\begin{equation}
    v = \mathcal{V}(h_v), ~~~~ a = \mathcal{A}(h_a)
\end{equation}
These tokens are then used as prompts to the LLM along with the time representations. Following prior work~\citep{cheng2024videollama,li2024groundinggpt}, we use the fine-tuned vicuna 7B-v1.5~\citep{chiang2023vicuna} as our LLM to generate the final text responses. 
Next, we discuss how to incorporate time into our model.
\vspace{-0.5em}
\subsection{Time Alignment between Audio and Video}
\label{sec.time}

Existing multimodal LLMs rely on learnable positional embeddings to encode the order of frames, but they struggle to capture the absolute time elapsed between frames and lack a fine-grained, cross-modal understanding of audio and video. We propose two strategies to encode absolute and relative temporal information on video and audio tokens, called Interleaving Time Tokens (ITT) and Rotary Time Embeddings (\rote).
\vspace{-0.5em}

\noindent
\textbf{Interleaving Time Tokens (ITT).} In this approach, we interleave time tokens with the audio and the visual features. We allocate a budget of K learnable time tokens, zero-indexed by $k_i$, and assign a time token to an audio-visual feature with the following indexing function:
\begin{equation}
\label{eq.discrete_time}
k_i = \text{round}\left(\frac{\tau_i}{T} \cdot (K-1)\right) 
\end{equation}
where $\tau_i$ is a continuous timestamp in seconds, $T$ is the total duration of the video or audio in seconds, and $K$ is the total number of learnable time tokens. 

For a video $V$ with duration $T$ and video token embeddings $v_i$ where $i = 1 \cdots M$, each embedding is associated with a timestamp $\tau_i$ (\eg~0.5 seconds, 1.4 seconds, and so forth). We first use these timestamps to obtain the discrete time tokens, then we interleave them with the visual tokens $v_i$ obtained after the visual adaptor layers. Specifically, each visual token $v_i$ corresponds to a discrete time token indexed by $k_i$, as described in \cref{eq.discrete_time}. Hence, the interleaved visual sequence is given as $\bar{v}$= $\{v_1, <k_1>, v_2, <k_2> \cdots, <v_M>, <k_M>\}$. 

Similarly, for the given audio $A$ of duration $T$, we extract $N$ windows of length $w$ from the audio sequence such that for each window the time is represented as: $\tau_n = [n, n+w]~~~~\text{for}~~~~ n = 1, 2, \cdots, N$, where $n$ is the time in seconds. We then take the mean of the time windows, $\tau_n = \frac{n + (n+w)}{2}$. Then, we convert $\tau_n$ into discrete time token $k_n$ using \cref{eq.discrete_time} and interleave them with the audio tokens $a$ obtained from the audio adaptor layers. Hence, the interleaved audio sequence is represented as $\bar{a}$ = $\{a_1, <k_1>, a_2, <k_2> \cdots, <a_N>, <k_N>\}$. 
The final interleaved tokens $\bar{v}$ and $\bar{a}$ are then concatenated with the text instructions as prompts to the LLM, as shown in \cref{fig.omcat} on upper top right.

% \noindent
% \arushi{1. Add $\tau$ is in seconds. 2. Add how audio and video timestamps are extracted by generailizing $\tau$. 3. Discuss how}
\vspace{-0.5em}

\noindent
\textbf{Rotary Time Embeddings (\rote).} While we could use RoPE~\citep{su2024roformer} and avoid the extra context length cost introduced by ITT, RoPE would still lack the ability to capture the absolute time elapsed between frames, which is very important and crucial in scenarios with varying frame rates. To address these limitations, we propose an alternative strategy called \rote: a modified version of RoPE, where the rotation angles are determined by absolute timestamps in seconds instead of frame indices. \rote~takes inspiration from a real clock, where each handle rotates at distinct speeds, or ``frequencies". Similarly, in \rote~we rotate different dimensions in the visual and audio feature embeddings given their timestamp in seconds and the respective ``frequency" of that dimension. Our results in \cref{sec:experiments} show that \rote~achieves performance that is superior to the baselines. A visual representation of \rote~is shown in \cref{fig.intro_figure} on the lower right bottom.

In practice, while in rope the angle for rotation $\theta$ is defined by the temporal indexing of a token $\theta \gets - i \times 2\pi$,  \rote~is defined by the absolute time $\theta \gets -\tau_i \times 2\pi$. These temporally enriched features are then passed to the adaptor layers $\mathcal{V}(.)$ and $\mathcal{A}(.)$ to create visual tokens $v$ and audio tokens $a$ respectively. 
% We discuss details about how to create instruction prompts to query the LLM in \cref{sec.prompts}.
\vspace{-0.5em}

\begin{table*}[t]
    \small
    \centering
    \caption{List of datasets used for training \modelname. TS indicates if timestamps are available. ST refers to single-turn question answers. MT is the version with multi-turn dialogue.}
    \renewcommand*{\arraystretch}{1.0}
    \resizebox{\linewidth}{!}{
    \begin{tabular}{cclcr}
        \toprule
        Stage & Modality & Datasets & TS & \#(Modality, Text) \\ \toprule
        \multirow{6}{*}{\shortstack{Stage I\\Alignment Tuning}} & Image & LLaVA-Pretrain-595k \citep{liu2024visual} & \ding{55} & 558128 \\ 
        
        & Audio & WavCaps \citep{mei2024wavcaps} & \ding{55} &  403044 \\ 
        & Video & Valley-703K \citep{luo2023valley} & \ding{55} &  703000 \\
        & Video & VATEX \citep{wang2019vatex} & \ding{55} &  227250 \\
        & Audio-Video & VAST \citep{chen2023vast} & \ding{55} & 414602 \\
        & Audio-Video & VALOR \citep{chen2023valor} & \ding{55} & 16109 \\ \midrule
        
        \multirow{24}{*}{\shortstack{Stage II\\ Instruction\\Tuning}} 
         &Image&LLaVA-Tune~\citep{liu2024visual}& \ding{55} &624610\\\cline{2-5}
         &\multirow{5}{*}{Audio} & VGG Sound \citep{chen2020vggsound} & \ding{55} & 5157 \\
         &&AudioCaps \citep{kim2019audiocaps} & \ding{55} & 49838 \\
         &&MusicCaps \citep{agostinelli2023musiclm} & \ding{55} & 2858 \\
         &&Clotho \citep{drossos2020clotho} & \ding{55} & 3938 \\
        &&Audioset-Strong \citep{hershey2021benefit} & \ding{51} & 431131 \\
        \cline{2-5}
       
        & \multirow{12}{*}{Video}  & VideoInstruct 100K \citep{maaz2023video}& \ding{55} & 98145 \\ 
        &&VideoChatGPT \citep{maaz2023video}& \ding{55} & 100010\\
        &&WebVidQA \citep{yang2022learning}& \ding{55} & 100000\\
        &&Valley-Instruct 65k \citep{luo2023valley}& \ding{55} & 64690\\
        &&VideoChat-Instruct \citep{li2023videochat}& \ding{55} & 6961\\
        &&Activitynet captions \citep{krishna2017dense}& \ding{55} & 7481\\
        &&NextQA \citep{xiao2021next}& \ding{55} & 34132\\
        &&DiDeMO \citep{anne2017localizing}& \ding{51} & 27935 \\
        &&Charades \citep{gao2017tall}& \ding{51} & 12408 \\
        &&ActivityNet-RTL \citep{huang2024lita}& \ding{51} & 33557 \\
        &&Youcook2 \citep{zhou2018towards}& \ding{51} & 8643 \\
        &&ActivityNet Dense captions\citep{krishna2017dense} & \ding{51} & 33212 \\\cline{2-5}
        & \multirow{6}{*}{Audio-Video}  & Macaw Instruct \citep{lyu2023macaw}& \ding{55} & 50656 \\
        
        &&AVQA \citep{yang2022avqa}& \ding{55} & 40425\\
        &&Music-AVQA \citep{li2022learning}& \ding{55} & 25854\\
        &&UnAV-100 \citep{geng2023dense}& \ding{51} & 10358\\
        &&\texttt{OCTAV-ST} (Ours) & \ding{51} & 127507\\\midrule
    \multirow{3}{*}{\shortstack{Stage III\\ Multi-turn Instruction\\Tuning}} 
         &\multirow{3}{*}{Audio-Video}&AVSD \citep{alamri2019audio}& \ding{55} & 159700\\
         &&UnAV-100-MT (Ours)& \ding{51} & 94916 \\
         &&\texttt{OCTAV-MT} (Ours)& \ding{51} & 86000 \\
    \bottomrule
    \end{tabular}
    }
    \label{tab:our_datasets}
\end{table*}
% \vspace{-0.5em}

\subsection{Instruction Prompts}
\label{sec.prompts}
\vspace{-0.5em}

In this section, we explain how video and audio tokens are combined with text prompts. The prompt format is as follows:

\text{User}: $<system~prompt>$ \text{Question} $<vi\_start>$ $<vi\_patch>$ $<vi\_end>$ $<so\_start>$ $<so\_patch>$ $<so\_end>$ $<vis\_start>$ $<vi\_patch>$ $<so\_patch>$ $<vis\_end>$ \text{Assistant:} 

Here, $<system~prompt>$ represents a guiding system message, following Vicuna-7B~\citep{chiang2023vicuna}. Visual and audio markers are included through tokens like $<vi\_start>$/$<vi\_end>$ for video and $<so\_start>$/$<so\_end>$ for audio. Video tokens ($<vi\_patch>$) encode visual information, and audio tokens ($<so\_patch>$) handle sound data. It is important to note that these individual video and audio markers are activated only when modality-specific data (video or audio) is present. For joint audio-video data, $<vis\_start>$/$<vis\_end>$ marks the boundaries, encoding both audio and video tokens, deactivating the individual representations.

\subsection{Training Strategy}
\label{sec.multistage}
% \vspace{-0.5em}
% In \cref{tab:our_datasets}, we discuss the datasets used in training the different stages of \modelname.

\noindent
\textbf{Stage I: Alignment Tuning Stage. }
% \karan{maybe we state in this stage we train our video and audio adapters... ? because currrent continuity is broken }
In this stage, we train the visual and audio adaptor layers and freeze the parameters of the pre-trained visual and audio encoders as well as the LLM, as shown in \cref{fig.omcat}. By doing so, the model can focus on learning robust features for the adaptor layers, which play a key role in bridging the gap between the raw audio-visual inputs and the semantic representations of the LLM.

% \vspace{-1em}

\cref{tab:our_datasets} lists the image-text pairs~\citep{liu2024visual}, video-text pairs~\citep{luo2023valley,wang2019vatex}, and audio-text pairs~\citep{mei2024wavcaps} that were used to train the visual and audio adaptor layers such that the visual and audio representations are ``aligned" with their corresponding textual description. In addition to these individual modalities, we also incorporate joint audio-video-text paired data~\citep{chen2023vast,chen2023valor} to simultaneously train both the audio and visual adaptor layers. In total, we approximately use $\sim$2.3M training data. This joint training process helps the model develop a deeper understanding of the relationships between the audio and visual modalities, improving the model's ability to handle multimodal data. 

% To ensure that the training focuses on aligning the audio and visual modalities with the embedding space of the language model (LLM), we freeze the parameters of the pre-trained visual and audio encoders as well as the LLM backbone \karan { LLM Backbone ???  just LLM  ?} during this phase. By doing so, the model can concentrate on refining the adaptor layers, which are responsible for bridging the gap between the raw audio-visual inputs and the text representations of the LLM, thus improving the overall multimodal understanding. 
% \karan{reword: By doing so, the model can focus on learning robust features for the adaptor layers, which play a key role in bridging the gap between the raw audio-visual inputs and the semantic representations of the LLM, thereby enhancing overall multimodal understanding.}
\vspace{-0.5em}

\noindent
\textbf{Stage II: Instruction Tuning Stage. }
Following the ``alignment" of modality features in Stage I, we proceed to train \modelname using a diverse and high-quality collection of multimodal data ($\sim$2.8M). This includes image-text, video-text, audio-text, and audio-video-text datasets that are carefully curated to prepare the model for a wide range of tasks involving video and audio. These tasks include fine-grained timestamped comprehension as well as cross-modal understanding, enabling the model to perform effectively across multiple input types. A comprehensive overview of the datasets used in this training phase is provided in \cref{tab:our_datasets}.
During this training stage, we freeze the parameters of both the visual and audio encoders. We only fine-tune the visual and audio adaptor layers, along with the large language model (LLM), allowing these components to be further optimized to handle multimodal tasks.

\vspace{-0.5em}

\noindent
\textbf{Stage III: Multi-Turn Instruction Tuning Stage.} In the third and final stage, our main focus is to enhance the capabilities of \modelname on multi-turn question answering in complex audio-visual scenarios. To achieve this, we fine-tune our model on multi-turn datasets, including our proposed \texttt{OCTAV-MT}, UnAV-100-MT, and AVSD~\citep{alamri2019audio}, a dataset for audio-visual dialog. Detailed statistics of these datasets are shown in \cref{tab:our_datasets}. Overall, we use $\sim$340k training data during this stage. In this stage as well, the video encoder and the audio encoder remain frozen while we optimize the audio/video adaptor layers, along with the LLM. 

% \karan{refer to table with OCTAV-MT?}
% \arushi{add comparison to training samples in existing models}

% \multirow{9}{*}{\shortstack{Stage-3\\Instruction tuning}} & 
%          Audio & Clotho & \ding{55} & 3938 \\\cline{2-4}
%         & \multirow{3}{*}{Video}  & Valley-Instruct 65K& \ding{55} & 64690 \\ 
%         &&VideoChat-Instruct 11K & \ding{55} & 40812 \\
%         &&ActivityNet Captions& \ding{55} & 7481\\\cline{2-4}
%         & \multirow{5}{*}{Audio-Video}  & Macaw Instruct & \ding{55} & 50656 \\ 
%         % &&Mixed UnAV-100 (Ours) &  \\
%         &&Mixed QueryD (Ours) & \ding{55} & 16800 \\
%         &&Mixed ActivityNet (Ours) & \ding{55} & 15986 \\
%         &&Mixed Youcook2 (Ours) & \ding{55} & 9246 \\\bottomrule

%% file: experiments.tex
% \arushi{note to self: also emphasize on how we curate different timestamped and non-timestamped data for training }
\vspace{-1em}

\noindent
\textbf{Datasets. }To evaluate the capabilities of \modelname on general multimodal understanding, we evaluate our method on audio-visual understanding benchmarks. Specifically, we evaluate on the AVSD dataset~\citep{alamri2019audio} which is a dataset for audio-visual scene aware dialog, Music-AVQA dataset~\citep{li2022learning} that has audio-visual question answering for the music domain and AVQA dataset~\citep{yang2022avqa} which has general questions about audio and visual modalities. 

Furthermore, to evaluate whether \modelname outperforms in temporal tasks, we measure the performance of our model on temporal video grounding benchmark, Charades-STA~\citep{gao2017tall}. This dataset is widely used in prior works~\citep{cheng2024videollama,li2024groundinggpt,ren2024timechat} as a benchmark for temporal understanding. 

Finally, we benchmark \modelname on the evaluation subset of \texttt{OCTAV-ST}, \texttt{OCTAV-MT} and \texttt{UnAV-100-MT} datasets. These tasks require fine-grained temporal understanding, cross-correlation between the audio and visual modalities and hence are a good measure to evaluate the capabilities of \modelname. 
% \label{sec.imp}
\vspace{-0.5em}

\noindent
\textbf{Evaluation metrics. }Following prior work~\citep{cheng2024videollama,li2024groundinggpt,tang2024avicuna}, we use GPT-4~\citep{achiam2023gpt} to evaluate the answers predicted by the model by comparing against the correct answers, with a score of 0 to 5 indicating the accuracy. Besides Charades-STA where we use Recall@1 at Intersection over Union (IoU) thresholds of 0.5 and 0.7, we use the GPT accuracy everywhere else. 
\vspace{-0.5em}

\noindent
\textbf{Architecture. }We use the pre-trained CLIP visual encoder ViT-L/14~\citep{radford2021learning} to extract video/image features. For the audio encoder, we use the pre-trained ImageBind~\citep{girdhar2023imagebind} model. Similar to previous work, for the video and audio adaptors, we use the Q-former which has the same architecture as the Q-Former in BLIP-2~\citep{li2023blip}. However, to maintain the temporal consistency of video and audio frames in the ITT setup, we replace the Q-Former adaptor layers with 2-layer transformer blocks with self-attention~\citep{vaswani2017attention}. During both training and inference, we sample 64 frames from the video and we extract five 3-second windows for the audio. The audio is resampled to 16KHz sampling rate and converted into spectrograms to be consistent with the input to the ImageBind model~\citep{girdhar2023imagebind}. We use 100 as the value of $K$, the learnable time tokens in \cref{sec.time}.
\vspace{-0.5em}

\noindent
\textbf{Training details. }During both the pre-training and fine-tuning stages, we train the model for one epoch on 8 NVIDIA A-100 GPUs. For the pre-training stage, we set the batch size of 64, learning rate of 1e-3 with a cosine learning decay and a warm-up period. In the fine-tuning stages, we set the batch size to 32, learning rate to 2e-5 with a cosine learning decay and a warm-up period and gradient accumulation to 2. Further details about training are given in \cref{sec.training_details}.
\vspace{0.5em}
% for debugging pdf compilation
\begin{table*}[!h]
    \small
    \centering
    \caption{Evaluation results for \modelname and other state-of-the-art models on AVQA tasks~\citep{yang2022avqa,alamri2019audio,li2022learning}, Charades-STA~\citep{gao2017tall} and our proposed \texttt{OCTAV-ST} dataset. While ${\dagger}$ describes results from models fine-tuned on the training set of those datasets, results in parentheses are zero-shot.}
    \renewcommand*{\arraystretch}{1.0}
    \resizebox{\linewidth}{!}{
        \begin{tabular}{l|c|ccc|cc|cc}
            \toprule
            \textbf{Method}  & \textbf{Time}  & \multicolumn{3}{c|}{\textbf{Accuracy}}& \textbf{R@1(IoU=0.5)} & \textbf{R@1(IoU=0.7)} & \multicolumn{2}{c}{{\textbf{Accuracy}}} \\ 
            & & \textbf{AVSD} & \textbf{Music-AVQA} & \textbf{AVQA} & \multicolumn{2}{c|}{\textbf{Charades-STA}} & \shortstack{\texttt{OCTAV-ST} \\ \textbf{Youcook2}} & \shortstack{\texttt{OCTAV-ST} \\ \textbf{ActivityNet}}\\ 
           \toprule
            PandaGPT~\citep{su2023pandagpt}&\ding{55}&26.1$^{\dagger}$&33.7& 79.8$^{\dagger}$ &-&-&x \\
            Video LLaMA~\citep{cheng2024videollama} &\ding{55}&36.7$^{\dagger}$&36.6& 81.0$^{\dagger}$ &3.8&0.9&x \\
            MacawLLM~\citep{lyu2023macaw} &\ding{55}&34.3$^{\dagger}$&31.8& 78.7$^{\dagger}$ & -&-&x \\
            AVLLM~\citep{shu2023audio} &\ding{55}&52.6$^{\dagger}$&45.2&- & -&-&x \\
            % CAT~\citep{} &\ding{55}&-&48.6& -& -&-&x \\
            % MEERKAT~\citep{} &\ding{55}&- &79.2$^{\dagger}$ & 87.1 &-&-&- & x \\
            AVicuna~\citep{tang2024avicuna}&\ding{51}&53.1$^{\dagger}$&49.6&- & -&-&-&- \\
            Video LLaMA 2~\citep{zhang2023video} &\ding{55}&\textbf{53.3}$^{\dagger}$&73.6$^{\dagger}$ & &-&-&9.14 & 10.55 \\
            
            GroundingGPT~\citep{li2024groundinggpt}&\ding{51}&-&-& -&29.6$^{\dagger}$&11.9$^{\dagger}$& 1.20$^{\dagger}$(3.87)& 1.57$^{\dagger}$(7.6)\\\midrule
            \rowcolor[rgb]{1.0, 0.8, 0.8}
            \modelname (RoTE)&\ding{51}& 49.4 $^{\dagger}$ & \textbf{73.8$^{\dagger}$(51.2)}& \textbf{90.2}$^{\dagger}$ &\textbf{32.3$^{\dagger}$}&\textbf{15.9$^{\dagger}$} & \textbf{16.9$^{\dagger}$(9.9)}&\textbf{19.0$^{\dagger}$(11.2)} \\
            \bottomrule
        \end{tabular}
    }
    \label{tab:main_results}
\end{table*}
\vspace{-1em}

\subsection{Quantitative Results}

\noindent
\textbf{Comparison to state-of-the-art. }
We follow previous work~\citep{cheng2024videollama,zhang2023video,shu2023audio} to evaluate \modelname on three audio-video understanding benchmarks. Based on the GPT-assisted evaluation scores in \cref{tab:main_results}, our model surpasses the most recent and relevant models on all benchmarks. While on Music-AVQA we achieve 51.2\% accuracy in the zero-shot setting and 73.8\% in the fine-tuned setting, outperforming SOTA models, on AVQA dataset we significantly outperform other models. We believe our competitive but relatively lower scores on AVSD comes from a difference in data distribution during the final training stage.

To evaluate temporal understanding in videos, we evaluate \modelname Charades-STA, an established benchmark for this task. We outperform GroundingGPT~\citep{li2024groundinggpt} on Recall@1 at IoU threshold of 0.5 and 0.7. This result shows that our method can also perform temporal understanding in the video domain. 

Finally, we present results on the single-turn version of our proposed \texttt{OCTAV} benchmark, \texttt{OCTAV-ST}. We evaluated VideoLLaMA2~\citep{zhang2023video} in a zero-shot setting on this dataset and fine-tuned GroundingGPT~\citep{li2024groundinggpt} on the \texttt{OCTAV-ST} training set for a fair comparison. As shown in \cref{tab:main_results}, our method outperforms all the above two methods in both the zero-shot (results in parantheses) and fine-tuned settings. These results confirm \modelname's ability to jointly learn cross-modal and temporal understanding from both video and audio data.
\begin{table*}[!h]
    \small
    \centering
    \vspace{0.2em}
    \caption{Results of different variations of \modelname (RoPE, ITT and \rote) on the \texttt{OCTAV-MT} benchmark and the UnAV-100-MT dataset.}
    \renewcommand*{\arraystretch}{1.0}
    \resizebox{\linewidth}{!}{
    \begin{tabular}{l|ccc}
        \toprule
        \textbf{Method}  & \multicolumn{3}{c}{\textbf{Accuracy}}  \\ 
        & \texttt{OCTAV-MT}-\textbf{Youcook2} & \texttt{OCTAV-MT}-\textbf{ActivityNet} & \textbf{UnAV-100-MT}\\ 
       \toprule
     
        GroundingGPT~\citep{li2024groundinggpt}       & 0.13            & 0.07  & 13.2 \\\midrule
        \modelname (RoPE)  & 3.3  & 2.4 & 15.7 \\
        \modelname (ITT)   & 3.1     & 4.1 & 16.6 \\
        \rowcolor[rgb]{1.0, 0.8, 0.8}
        \modelname (\rote) & \textbf{3.7} & \textbf{5.6} & \textbf{19.9} \\
        \bottomrule
    \end{tabular}}
    \label{tab:octav_mt}
\end{table*}

% \vspace{-0.5em}

\noindent
\textbf{Comparison on the \texttt{OCTAV-MT} benchmark. }
In \cref{tab:octav_mt}, we highlight the performance of \modelname on the \texttt{OCTAV-MT} benchmark, which involves multi-turn question-answer pairs for videos with multiple sound events. All models in \cref{tab:octav_mt} are fine-tuned on the proposed \texttt{OCTAV-MT} benchmark. Our model, \modelname with \rote, significantly outperforms the baselines—ITT, RoPE, and GroundingGPT~\citep{li2024groundinggpt}—on this dataset. Moreover, it achieves substantial performance gains on the UnAV-100-MT dataset, a dataset with in-the-wild/natural audio-visual events (\eg~\cref{fig.octav_mt_unav}).

\modelname with \rote~efficiently integrates time representations with minimal computational cost, ensuring precise cross-modal alignment between audio and video. While these improvements over the baselines are considerable, there is still ample room for further enhancement in this area. The \texttt{OCTAV-MT} benchmark paves the way for the development of advanced multimodal models with stronger cross-modal grounding capabilities.

\vspace{0.5em}

\begin{table*}[!h]
    \small
    \centering
    \caption{Effect of applying various time embeddings--RoPE, ITT and \rote~to \modelname on all benchmarks.}
    \renewcommand*{\arraystretch}{1.0}
    \resizebox{\linewidth}{!}{
    \begin{tabular}{l|ccc|cc|cc}
        \toprule
        \textbf{{\shortstack{Time\\Encoding}}}  & \multicolumn{3}{c|}{\textbf{Accuracy}} & \textbf{R@1(IoU=0.5)} & \textbf{R@1(IoU=0.7)} & \multicolumn{2}{c}{{\textbf{Accuracy}}} \\ 
        & \textbf{AVSD} & \textbf{Music-AVQA} & \textbf{AVQA} & \multicolumn{2}{c|}{\textbf{Charades-STA}} & \textbf{\texttt{OCTAV-ST}-Youcook2} & \textbf{\texttt{OCTAV-ST}-ActivityNet} \\ 
       \toprule
        RoPE  & 45.9 & 71.2 & 88.2 & 30.7 & 16.1 & 13.3 & 16.5 \\
        ITT   & 47.3 & 69.7 & 82.1 & \textbf{32.5} & \textbf{16.7} & 16.5 & \textbf{19.2} \\
        \rowcolor[rgb]{1.0, 0.8, 0.8}
        \rote & \textbf{49.4} & \textbf{73.8} & \textbf{90.2} &32.3 & 15.9 & \textbf{16.9}& 19.0 \\\bottomrule
    \end{tabular}}
    \label{tab:ablation_main}
\end{table*}

\begin{table*}[!h]
    \small
    \centering 
    \caption{Effect of alignment tuning data on the overall performance. LP denotes LLaVA-Pretrain-595k~\citep{liu2024visual}, WC denotes WavCaps~\citep{mei2024wavcaps} and, V denotes Valley-703K~\citep{luo2023valley}.}
    \renewcommand*{\arraystretch}{1.0}
    \resizebox{0.9\linewidth}{!}{
    \begin{tabular}{c|ccc}
        \toprule
        \textbf{Ablation}  & \textbf{Music-AVQA} & \textbf{Charades-STA (R@1,IoU-0.5)} & \textbf{\texttt{OCTAV-ST}-Youcook2}   \\ \toprule
     % \modelname w/ Type 1 rep &  x &x&x \\
      \modelname w/ only LP,WC,V   & 50.6 & 26.9 & 4.97   \\ \midrule
      \rowcolor[rgb]{1.0, 0.8, 0.8}
      Ours & \textbf{51.2} & \textbf{32.3} & \textbf{16.9} \\
         \bottomrule    
        \end{tabular}}
    \label{tab:stage1_results}
\end{table*}

% \noindent
\subsection{Ablation study}
\vspace{-0.5em}

\textbf{How does time embedding affect \modelname?} In \cref{tab:ablation_main}, we evaluate three different time embedding approaches, including RoPE~\citep{su2024roformer}, and our proposed approaches ITT and \rote. On the AVQA benchmark, \rote~consistently outperforms the baselines by a large margin, demonstrating its strong capability not only on temporal and cross-modal tasks but also in handling coarse-grained question answering. 

For the temporal understanding task on Charades-STA, ITT performs slightly better than \rote~at both IoU thresholds (0.5 and 0.7). On the \texttt{OCTAV-ST} benchmark, YouCook2 and ActivityNet, ITT and \rote~show nearly equivalent performance. We believe ITT's competitive results stem from its explicit time embedding through time tokens. However, given ITT’s increased context length and its weaker performance on AVQA tasks, \rote{} is the more effective and efficient choice overall.

\vspace{-0.5em}
\noindent
\textbf{What is the effect of pre-training data on \modelname? }Furthermore, we investigate the impact of pre-training data on the final model performance, particularly during the alignment tuning stage (Stage I). This stage is crucial for establishing the model's capacity to ``align" information across different modalities, such as image, video, and audio, with text. To examine the role of joint multimodal data, we conduct an ablation study where we modify the training data by excluding the audio-video-text paired data~\citep{chen2023vast,chen2023valor} while retaining image-text~\citep{liu2024visual}, video-text~\citep{luo2023valley,wang2019vatex}, and audio-text pairs~\citep{mei2024wavcaps}. 

Our results in \cref{tab:stage1_results} indicate a noticeable decline in performance across all tasks when the model is trained without audio-video-text data. This demonstrates the critical importance of joint multimodal data in achieving robust cross-modal alignment. We hypothesize that without data that directly links audio, video, and text, the model struggles to accurately capture the intricate relationships between these modalities, leading to suboptimal performance in tasks requiring fine-grained multimodal understanding.

% \begin{table*}[!h]
%     \small
%     \centering
%     \caption{\arushi{$^{\dagger}$ means the model is fine-tuned on that dataset otherwise the results are zero shot}}
%     \renewcommand*{\arraystretch}{1.0}
%     \resizebox{\linewidth}{!}{
%     \begin{tabular}{c|ccc|ccc}
%         \toprule
%         \textbf{Method}  & \multicolumn{3}{c|}{\textbf{Accuracy}} & \multicolumn{3}{c}{\textbf{R@1(IoU=0.5)}} \\ 
%         & \shortstack{{\texttt{OCTAV-MT}-\\\textbf{Youcook2}}} & \shortstack{{\texttt{OCTAV-MT}-\\\textbf{ActivityNet}}} & \textbf{UnAV-100-MT}& \shortstack{{\texttt{OCTAV-MT}-\\\textbf{Youcook2}}} & \shortstack{{\texttt{OCTAV-MT}-\\\textbf{ActivityNet}}} & \textbf{UnAV-100-MT}\\ 
%        \toprule
     
%         GroundingGPT~\citep{}&-&-& -&-&-&\\\midrule
%         \modelname (Vanilla) &-&gpt-avsd& dc-03&-&-& \\
%         \modelname (ITT) &-&gpt-mixed-activity& dc-02&-&-& \\
%         \modelname (RoTE) &-&gpt-activity& dc-01&-&-& \\
%         \bottomrule
%     \end{tabular}}
%     \label{tab:main_results}
% \end{table*}

% \noindent
% \textbf{Effect of token representations. }

% 1) Type 1: $<vi\_start>$ $<vi\_patch>$ $<vi\_end>$ $<so\_start>$ $<so\_patch>$ $<so\_end>$
% \\
% 2)Type 2: $<vi\_start>$ $<vi\_patch>$ $<vi\_end>$ $<so\_start>$ $<so\_patch>$ $<so\_end>$ $<vis\_start>$ $<vi\_patch>$ $<so\_patch>$ $<vis\_end>$

% \begin{figure*}[h]
%     \centering
%     \includegraphics[width=\linewidth]{figures/performance_comparison.pdf}
%     \vspace{-0.5em}
%     \caption{}
%     \label{fig.ablation_training_data}
% \end{figure*}

\begin{figure*}[!h]
    \centering
    \includegraphics[width=\linewidth]{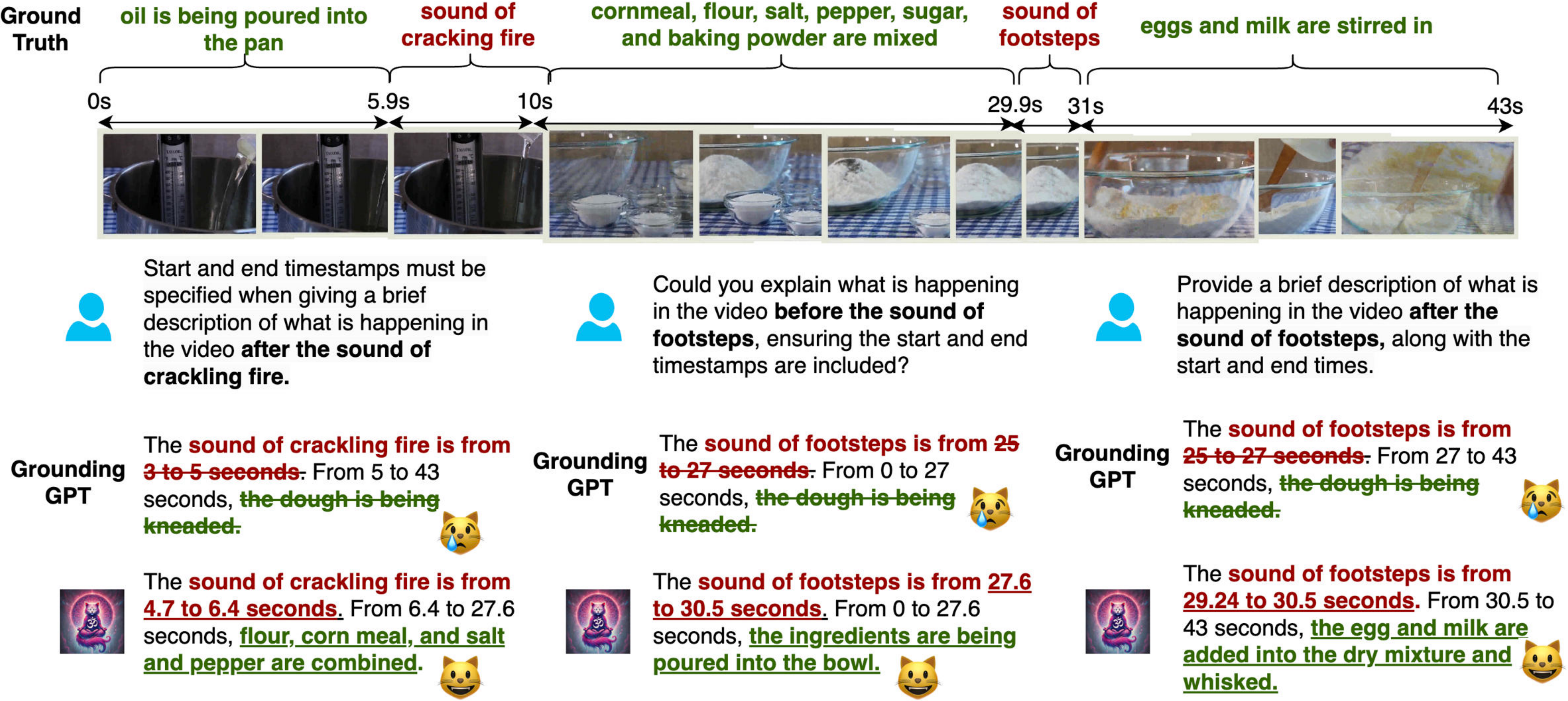}
    \vspace{-0.5em}
    \caption{Qualitative comparison of \modelname with GroundingGPT on the \texttt{OCTAV-MT} dataset.}
    \label{fig.qual_results}
\end{figure*}

\vspace{-0.7em}

\subsection{Qualitative Results}
\vspace{-0.5em}
% \paragraph{Qualitative Results}

In \cref{fig.qual_results}, we showcase the qualitative performance of our method on the YouCook2 subset of the \texttt{OCTAV-MT} benchmark. GroundingGPT inaccurately predicts a uniform activity of \textit{dough being kneaded}, failing to capture the nuanced transitions in events triggered by sound cues. In contrast, our model successfully isolates specific events and accurately associates them with their corresponding timestamps based on the sound events.
For instance, our model correctly identifies the activity following the \textit{sound of cracking fire} (around 6.4 to 27.6 seconds), predicting that \textit{flour, cornmeal, and salt and pepper are combined}. This aligns closely with the ground truth, which describes the activity as \textit{cornmeal, flour, salt, pepper, sugar, and baking powder being mixed}. While \modelname omits some ingredients, it still recognizes the correct activity—unlike GroundingGPT, which mistakenly predicts \textit{dough being kneaded}.

Similarly, \modelname accurately predicts that \textit{egg and milk are added into the dry mixture and whisked} following the \textit{sound of footsteps} (from 29.2 to 30.5 seconds). However, when asked what occurs before the sound of footsteps, the model correctly predicts the activity as \textit{ingredients being mixed in the bowl}, though the prediction does not perfectly match the ground truth.

% \vspace{-0.5em}

%% file: discussion.tex
\vspace{-1em}

In this paper, we addressed the limitations of multimodal large language models in fine-grained, cross-modal temporal understanding by introducing the \texttt{OCTAV} dataset and the \modelname model. \texttt{OCTAV} focuses on event transitions across audio and video, promoting deeper temporal alignment and cross-modal understanding. \modelname, enhanced with \rote{} embeddings, effectively grounds temporal information across modalities, leading to superior performance on Audio-Visual Question Answering (AVQA) tasks and the \texttt{OCTAV} benchmark. Our approach sets a new standard for multimodal AI, advancing cross-modal and temporal reasoning capabilities for future research.

% \arushi{add limitations}

%% file: appendix.tex
\section{Demo Page link}\label{app:website}
The link to our demo page is \url{https://om-cat.github.io/}.

\section{Prompts for generating \texttt{OCTAV} dataset}
In this section, we discuss further details about generating our proposed dataset. 
\subsection{Prompts for \texttt{OCTAV-ST} dataset}
\label{sec.octav_st_prompt}
Below we show the prompts used to generate question-answer pairs for the video conditioned on a single audio event~\ie~\texttt{OCTAV-ST} dataset. 
\begin{table}[h]
    \centering
    \begin{tabular}{|p{13.5cm}|}
        % Drawing a rectangle
        \hline
% \begin{tcolorbox}[boxsep=0.5pt,left=2pt,right=2pt,top=1pt,bottom=1pt,colback=white!5!white,colframe=gray!75!black]
You are an AI assistant that can analyze a video.
You receive timestamped video and audio captions with start time and end times describing the video you are observing. 
Based on these audio and video captions, create 2 question and answer pairs where a question is asked by the person (the user) and the answer is given by you (the assistant) about the events in the video/audio. 
Here are some additional requirements about the generated question-answer pairs:\\
1. The question asked by the user should be from the audio caption and the answer given by the assistant should be from the video caption before or after that timestamp in question.\\
2. Only describe what you are certain about, and avoid providing descriptions that maybe ambiguous or inaccurate.\\
4. The number of words in the answer should not exceed 100 words. Keep it as concise as possible. You do not need to include everything in the answer.\\
Include timestamp information in the answers.\\
\\
Example 1:
\\
Timestamped video and audio captions: \\
``video caption 1": season the chicken on both sides with salt and pepper then cut it into pieces from 0.0 to 18,
``video caption 2": put the chicken pieces to a boiling pot of water cover it and let it cook from 20 to 22,
``audio caption": There is a sound of Trumpet from 18 to 20.

\\
QA: \\

User: What is happening in the video before the sound of trumpet? 
Assistant: The sound of trumpet is from [18.0, 20.0]. From [0.0, 18.0], the chicken is seasoned on both sides with salt and pepper then cut it into pieces.

User: What is happening in the video after the sound of trumpet? 
Assistant: The sound of trumpet is from [18.0, 20.0]. From [20.0, 22.0], the chicken pieces are put to a boiling pot of water, covered and then cooked.
\\\\
Based on the example above, design 2 question and answer pairs between the user and assistant for the example given below. 
\\
Format each QA pair in a single line as a JSON dictionary (key ``user" for question, and ``assistant" for answer). \\
\hline
\end{tabular}
% \label{table.prompt_octav}

\end{table}
% \end{tcolorbox}
\subsection{Prompts for \texttt{OCTAV-MT} dataset}
\label{sec.octav_mt_prompt}
Below we show the prompts used to generate question-answer pairs for the video conditioned on two audio events~\ie~\texttt{OCTAV-MT} dataset. 
\newpage

\begin{table}[h]
    \centering
    \begin{tabular}{|p{13.5cm}|}
        % Drawing a rectangle
        \hline
% \begin{tcolorbox}[boxsep=0.5pt,left=2pt,right=2pt,top=1pt,bottom=1pt,colback=white!5!white,colframe=gray!75!black]
You are an AI assistant that can analyze a video.
You receive timestamped video and audio captions with start time and end times describing the video you are observing. 
Based on these audio and video captions, create 4 question and answer pairs where a question is asked by the person (the user) and the answer is given by you (the assistant) about the events in the video/audio. 
You can ask clarification questions if the question asked by the user is not clear.
Here are some additional requirements about the generated question-answer pairs:\\
1. The question asked by the user can be from the audio caption or the video caption and the answer given by the assistant should be from the video caption before or after that timestamp in question.\\
2. Only describe what you are certain about, and avoid providing descriptions that maybe ambiguous or inaccurate.\\
4. The number of words in the answer should not exceed 100 words. Keep it as concise as possible. You do not need to include everything in the answer.\\
Include timestamp information in the answers.\\
\\\\
Example 1:
\\
Timestamped video and audio captions:\\
        
``video caption 1": season the chicken on both sides with salt and pepper then cut it into pieces from 0.0 to 18,
``video caption 2": put the chicken pieces to a boiling pot of water cover it and let it cook from 20 to 22,
``video caption 3": chop celery to small pieces chop cheese to cubes and chop ham also to the same size from 26 to 50,
``audio caption 1": There is a sound of laugh from 18 to 20,
``audio caption 2": There is a sound of laugh from 22 to 26.
\\
\\
QA: 

User: What is happening in the video after the sound of laugh? Answer with start and end timestamps.\\
Assistant: There are two sounds of laugh, one from [18.0, 20.0] and the other one from [22.0, 26.0]. Which laugh are you referring to?\\

User: I am referring to the laugh that happens after the chicken pieces are out to a boiling pot of water.\\
Assistant: Okay, so the laugh from [22.0, 26.0]. After this sound of laugh from [26.0, 50.0], celery is chopped to small pieces, cheese is chopped to cubes and ham is chopped also to the same size.\\

User: Thanks, what is happening in the video after the chicken is seasoned on both sides with salt and pepper. Answer with start and end timestamps.\\
Assistant: There is a sound of laugh from [18.0, 20.0] and from [20.0, 22.0], the chicken pieces are put ot a boiling pot of water, covered and cooked.\\

User: Thanks, what is happening in the video after the sound of bird chirping? Answer with start and end timestamps.\\
Assistant: Sorry, there is no sound of bird chirping. \\

% Example 2:

% Timestamped video and audio captions: 
      
% ``video caption 1": add 3 tbsp of vinegar from 0.0 to 12,
% ``video caption 2": add 1 beaten egg to the soup while stirring from 15 to 25,
% ``video caption 3": sprinkle with scallion greens from 28 to 32,
% ``audio caption 1": There is a sound of Bass guitar from 12 to 15.,
% ``audio caption 2": There is a sound of yawn from 25 to 28.

% \\
% \\
% QA: 

% User: What is happening in the video before the sound of bass guitar? Answer with start and end timestamps.\\
% Assistant: The sound of bass guitar is from [12.0, 15.0]. From [0.0, 12.0], 3 tbsp of vinegar is added.\\

% User: What is happening in the video after the sound of bass guitar and before the sound of yawn? Answer with start and end timestamps.\\
% Assistant: The sound of bass guitar is from [12.0, 15.0] and sound of yawn is from [25.0, 28.0]. From [15.0, 25.0], 1 beaten egg is added to the soup while stirring.\\

% User: What is happening in the video after the sound of yawn? Answer with start and end timestamps.\\
% Assistant: The sound of yawn is from [25.0, 28.0]. From [28.0, 32.0], scallion greens are sprinkled.\\

% User: Thanks, what is happening in the video after the sound of laugh? Answer with start and end timestamps.\\
% Assistant: Sorry, there is no sound of laugh in this video. \\
\\
Based on the above examples, design 4 question and answer pairs between the user and assistant for the example given below. \\

Format each QA pair in a single line as a JSON dictionary (key "user" for question, and "assistant" for answer, wrapped with { and }). \\
\hline
\end{tabular}
% \label{table.prompt_octav}

\end{table}
% \end{tcolorbox}

%Examples of OCTAV dataset from each subset
\subsection{List of instructions}
\label{sec.instructions}

Below, we show the diverse set of instructions that we use to replace the common instruction \textit{What is happening in the video} generated by the GPT model. The ${eventname}$ below is replaced by the anchored query such as \textit{after the sound of bird chirping}.
\newpage
\begin{table}[h]
    \centering
    \begin{tabular}{|p{13.5cm}|}
        % Drawing a rectangle
        \hline
% \begin{tcolorbox}[boxsep=0.5pt,left=2pt,right=2pt,top=1pt,bottom=1pt,colback=white!5!white,colframe=gray!75!black]
Start and end timestamps should be included while describing what ${event name}$ is.\\\\
Please include the start and end time when briefly describing what ${event name}$ entails.\\\\
Start and end timestamps are required while providing a brief description of what ${event name}$ involves.\\\\
Include the exact start and end times when describing what ${event name}$ refers to.\\\\
Ensure to mention the start and end timestamps when explaining what ${event name}$ covers.\\\\
With the start and end times, please provide a brief explanation of what ${event name}$ is.\\\\
Start and end timestamps should be given alongside a description of what ${event name}$ involves.\\\\
When describing what ${event name}$ is, include the exact start and end time information.\\\\
Include start and end time details when summarizing what ${event name}$ entails.\\\\
Start and end timestamps must be specified when giving a brief description of what ${event name}$ refers to.\\\\

Describe what ${event name}$ is with start and end timestamps.\\\\
Please briefly describe what ${event name}$ entails, including its exact start and end timestamps.\\\\
Provide a brief description of what ${event name}$ includes, along with the start and end times.\\\\
Give a short description of what ${event name}$ is, including the precise start and end time details.\\\\
Briefly explain what ${event name}$ involves, including its start and end timestamps.\\\\
Please summarize what ${event name}$ covers, specifying the start and end timestamps.\\\\
Give a brief explanation of what ${event name}$ is, making sure to include both the start and end times.\\\\
Could you describe what ${event name}$ refers to, including the exact start and end times?\\\\
Please provide a concise overview of what ${event name}$ involves, along with start and end time details.\\\\
Could you explain what ${event name}$ is, ensuring the start and end timestamps are included?\\\\
\hline
\end{tabular}
% \label{table.prompt_octav}

\end{table}
% \end{tcolorbox}

\subsection{Sound events}
\label{sec.sound_events}
In this section, we provide details about the datasets we used for adding sound to the curated chunked videos as discussed in \cref{sec:dataset}. Specifically, we use Urban Sound 8K~\citep{salamon2014dataset}, ESC-50~\citep{piczak2015esc}, FSD50K~\citep{fonseca2021fsd50k} and NonSpeech7K~\citep{rashid2023nonspeech7k} datasets. 

Urban Sound 8K~\citep{salamon2014dataset} is an audio dataset that contains urban sounds from 10 classes: air conditioner, car horn, children playing, dog bark, drilling, engine idling, gun shot, jackhammer, siren, and street music.

The ESC-50 dataset~\citep{piczak2015esc} consists of 5-second-long recordings organized into 50 semantical classes that can be categorized into 5 major categories of animals, natural soundscapes \& water sounds, human and non-speech sounds, interior/domestic sounds and exterior/urban noises.

FSD50K~~\citep{fonseca2021fsd50k} has 200 sound categories mainly produced by physical sound sources and production mechanisms, including human sounds, sounds of things, animals, natural sounds, musical instruments and more.

Nonspeech7k~\citep{rashid2023nonspeech7k} contains a diverse set of human non-speech sounds, such as the sounds of breathing, coughing, crying, laughing, screaming, sneezing, and yawning.

\section{Examples from the \texttt{OCTAV} dataset}
\label{sec.examples_octav}
\begin{figure*}[t]
    \centering
    \includegraphics[width=\linewidth]{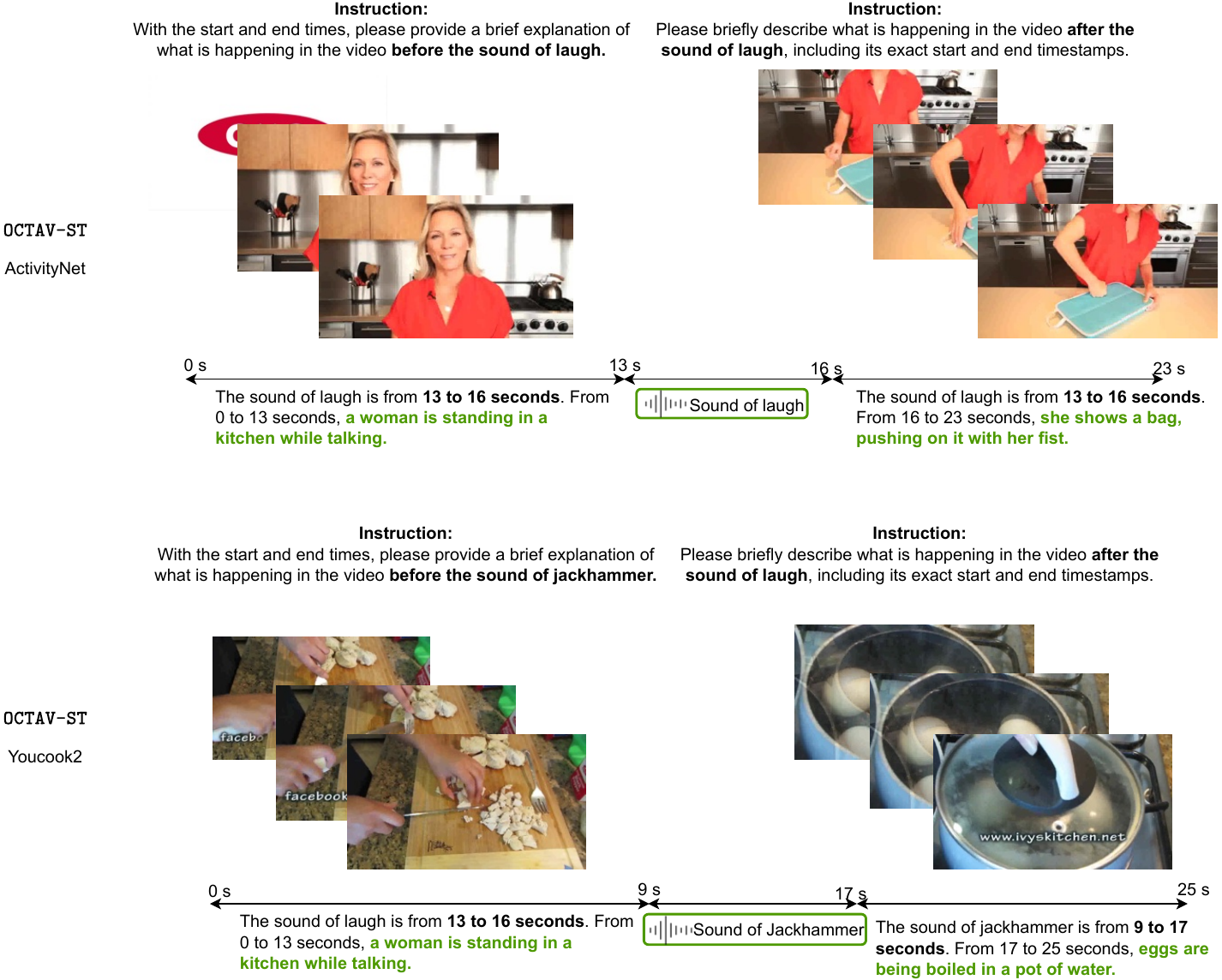}
    \vspace{0.5em}
    \caption{Question-answer pairs from the proposed \texttt{OCTAV-ST} dataset.}
    \vspace{2em}

    \label{fig.octav_st}
\end{figure*}

In \cref{fig.octav_st}, we show examples from the \texttt{OCTAV-ST} dataset. The top part of the figure shows an example from the ActivityNet subset and the bottom part shows an example from the Youcook2 subset of the dataset. These examples give an overview of how different event transitions are interwoven seamlessly with an audio event.

\begin{figure*}[!h]
    \centering
    \includegraphics[width=0.9\linewidth]{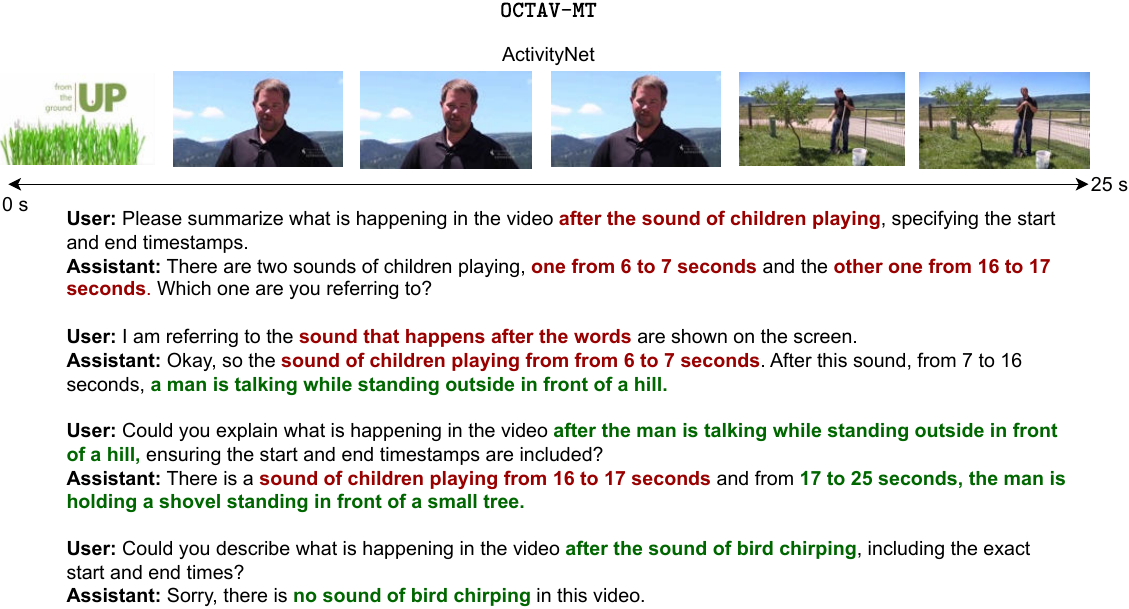}
        \vspace{0.5em}

    \caption{Multi-turn dialogue examples from the ActivityNet subset of our \texttt{OCTAV-MT} dataset.}
    \label{fig.octav_mt_anet}
\end{figure*}

\begin{figure*}[h]
    \centering
    \includegraphics[width=\linewidth]{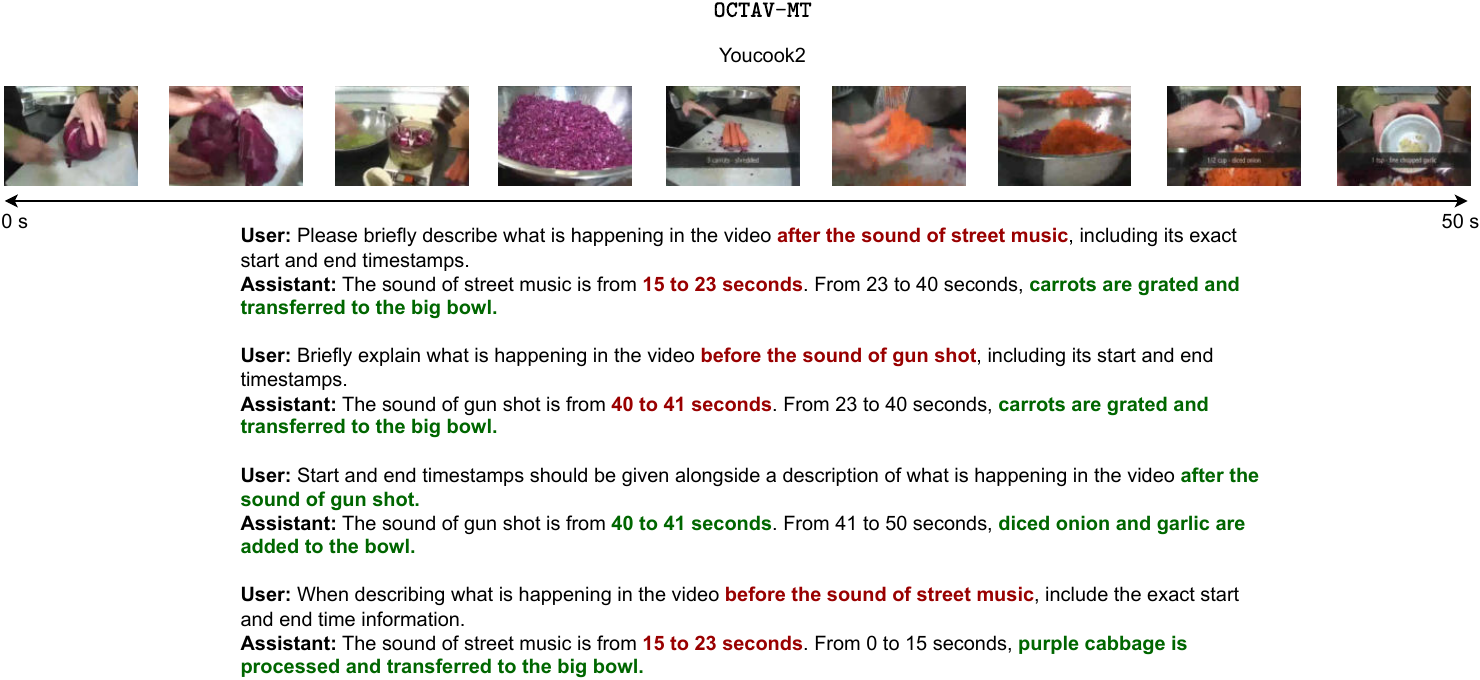}
    \vspace{0.5em}
    \caption{Multi-turn dialogue examples from the Youcook2 subset of our \texttt{OCTAV-MT} dataset.}
        \vspace{2em}

    \label{fig.octav_mt_youcook2}
\end{figure*}

In \cref{fig.octav_mt_anet} and \cref{fig.octav_mt_youcook2}, we show examples from the ActivityNet subset and the Youcook2 subset of the \texttt{OCTAV-MT} dataset respectively. These examples show the anchoring of transitioning video events on multiple sound events.

\begin{figure*}[ht]
    \centering
    \includegraphics[width=0.9\linewidth]{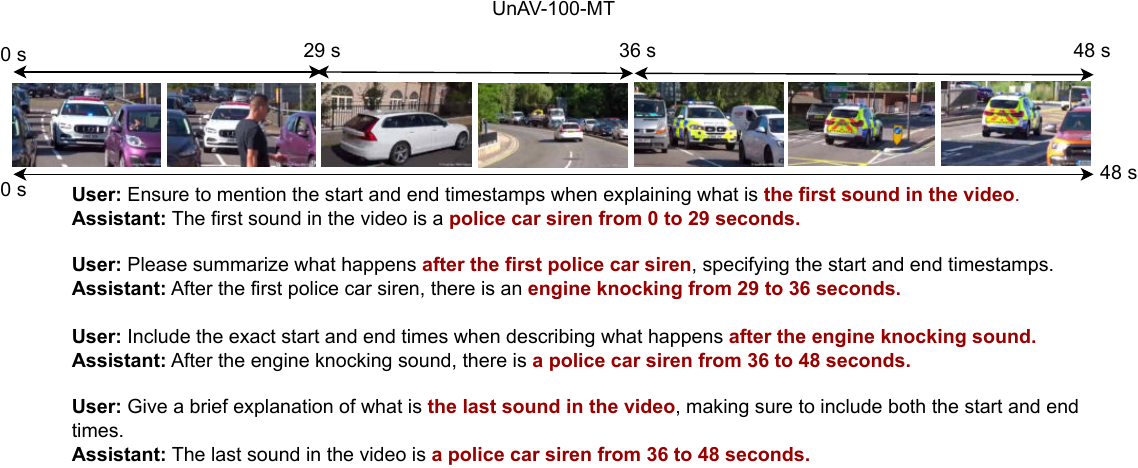}
    \vspace{1em}
    \caption{Multi-turn dialogue examples from the UnAV-100-MT dataset.}
    \vspace{4em}
    \label{fig.octav_mt_unav}
\end{figure*}

In \cref{fig.octav_mt_unav}, we show an example from the UnAV-100-MT dataset, which is the multi-turn version of the UnAV-100 dataset~\citep{geng2023dense}. We convert the audio-visual timestamped annotations from the UnAV-100 dataset into multi-turn question answers as shown in this example. This dataset acts as a benchmark for a real time setting of audio-visual scenarios.
\newpage


\newpage
\section{Training Details}
\label{sec.training_details}
The training settings for our model are spread across three stages, each with specific configurations.

Stage 1 utilizes a batch size of 64 and a learning rate of 1e-3, employing a cosine decay learning schedule. The warm-up ratio is set at 0.03, with no weight decay applied. This stage runs for 1 epoch with gradient accumulation of 1. Additionally, it employs the ZeRO2 optimization strategy in DeepSpeed~\citep{rasley2020deepspeed} and utilizes 8 A100 GPUs.

Stage 2 has a smaller batch size of 32 and reduces the learning rate to 2e-5. It follows the same warm-up ratio of 0.03 and applies no weight decay. Like Stage 1, this stage runs for 1 epoch but increases gradient accumulation to 2. The same DeepSpeed optimization and GPU configuration are used.

Stage 3 mirrors the settings of Stage 2, with a batch size of 32, a learning rate of 2e-5, and a warm-up ratio of 0.03. It also has no weight decay and runs for 1 epoch with gradient accumulation set to 2. We use the same DeepSpeed optimization and 8 A100 GPUs like the previous stages.

% \begin{table*}[!h]
%     \small
%     \centering
%     \caption{Comparison to existing state-of-the-art methods based on Pre-Training (PT) and SFT datasets }
%     \renewcommand*{\arraystretch}{1.0}
%     \resizebox{\linewidth}{!}{
%     \begin{tabular}{cccc}
%         \toprule
%         Method & PT datasets & SFT datasets & \#Total Pairs \\ \toprule
%         \multirow{3}{*}{Grounding GPT~\citep{}} & LCS-558K & DiDeMo, HiREST, Charades-STA & \\
%         & Valley-703K& VGGSS, Valley-Instruct-73k, Videochat-Instruct-11k \\ 
%         & WavCaps & Activitynet Captions, Clotho & \\ \midrule 
%         AVLLM~\citep{} & LCS-558K, Valley-770K & Curated (ACAV,WavCaps,VGGSound,WebVid2M)-260K & 1.6M \\ \midrule
%         CAT~\citep{} & WebVid-2.5M, WavCaps & AVInstruct, VideoChatGPT-100K & \\ \midrule
%         Video LLaMA 2~\citep{} & & & \\ \midrule
%         \multirow{2}{*}{AVicuna~\citep{}} & LCS-558K,AudioSet & PU-VALOR, InternVid, UnAV-100 & 1.1M\\
%         & AudioCap, Auto-CAD & VideoInstruct100K, ActivityNet Captions, DiDeMo & \\ \midrule
%         AAV-LLM (Ours)~\citep{} & & & 3.9M\\ \bottomrule    
%         \end{tabular}}
%     \label{tab:dataset_comparison}
% \end{table*}

\section{More Results}
In \cref{tab:video_results}, we showcase the zero-shot performance of our proposed \modelname model on the video understanding benchmarks MSRVTT-QA~\citep{xu2016msr}, MSVD-QA~\citep{chen2011collecting}, and ActivityNet-QA~\citep{yu2019activitynet}. Although our model's performance falls short compared to Video LLaMA 2~\citep{cheng2024videollama} and AVicuna~\citep{tang2024avicuna}, it remains competitive with other models in the field~\citep{li2024groundinggpt,zhang2023video,li2023videochat}. We attribute AVicuna's higher performance to its instruction tuning with ActivityNet captions~\citep{krishna2017dense} and its specialization in video understanding during the final training stage. Similarly, Video LLaMA 2~\citep{cheng2024videollama} is also an expert model, having been trained on a significantly larger video-text dataset throughout all training phases, unlike \modelname.

We further assess our method's effectiveness in audio understanding by evaluating it on the Clotho-AQA~\citep{lipping2022clotho} dataset, where \modelname achieves a score of 54.3\% in audio question answering. In comparison, the audio expert model Qwen-Audio~\citep{chu2023qwen} scores 57.9\%, while Video LLaMA 2 reaches 59.7\%. Our model demonstrates competitive performance on this benchmark; however, we believe that the extensive audio-text training data utilized by these two models contributes to their superior results. Moreover, we use Imagebind~\citep{girdhar2023imagebind} as our audio encoder whereas these models use a far more superior audio encoder pre-trained on a large-scale audio-text data unlike Imagebind~\citep{girdhar2023imagebind}. It is worth noting that this aspect was beyond the scope of our work, which primarily focuses on temporal and cross-modal understanding of audio and video.

% \vspace{1em}
\begin{table*}[!h]
    \small
    \centering
    \caption{Performance comparison on video understanding benchmarks. $\dagger$ means specialized model and $*$ means trained on a much larger dataset.}
    \renewcommand*{\arraystretch}{1.0}
    \resizebox{0.9\linewidth}{!}{
    \begin{tabular}{l|c|cccc}
        \toprule
        \textbf{Method}  & \textbf{Modality} & \textbf{MSRVTT-QA} & \textbf{MSVD-QA} & \textbf{ActivityNet-QA}  \\ \toprule
        VideoChat~\citep{li2023videochat} & Video &45.0 & 56.3 & 26.5 \\
        Video-ChatGPT~\citep{maaz2023video} & Video &49.3 &64.9 &35.2 \\
        Valley~\citep{luo2023valley} & Video & 45.7 & 65.4 & 42.9  \\
        Video-LLaMA~\citep{zhang2023video} &Video &29.6 & 51.6 & 12.4  \\
        PandaGPT~\citep{su2023pandagpt} &Video, Audio &23.7 & 46.7 & 11.2  \\
        MacawLLM~\citep{lyu2023macaw} &Video, Audio  & 25.5 & 42.1 & 14.5 \\
        AVLLM~\citep{shu2023audio} &Video, Audio & 53.7 & 67.3 & 47.2  \\
        GroundingGPT~\citep{li2024groundinggpt} &Video, Audio & 51.6 & 67.8 & 44.7 \\\midrule
        % CAT~\citep{} &Video, Audio & 62.1 & - & 50.2\\\midrule
        AVicuna$^\dagger$~\citep{tang2024avicuna} &Video, Audio &\textbf{59.7} & 70.2 & \textbf{53.0} \\
        Video LLaMA 2$^*$~\citep{cheng2024videollama} &Video, Audio & 53.9 & \textbf{71.7}& 49.9 \\
        \midrule
        \modelname (RoPE~\citep{su2024roformer})&Video, Audio & 49.3 & 63.2 & 41.9 \\
        \modelname (ITT)&Video, Audio & 51.1 & 65.1 & 43.9 \\
        \modelname (\rote)&Video, Audio & 51.2 & 67.8 & 46.6 \\
    
         \bottomrule    
        \end{tabular}}
    \label{tab:video_results}
\end{table*}

% \begin{figure*}[h]
%     \centering
%     \includegraphics[width=\linewidth]{figures/eg_mt_activity.pdf}
%     \vspace{-0.5em}
%     \caption{Dialogue example from the proposed OCTAV Multi-Turn dataset}
%     \label{fig.octav_mt_activity}
% \end{figure*}

% \begin{table*}[!h]
%     \small
%     \centering 
%     \caption{ }
%     \renewcommand*{\arraystretch}{1.0}
%     \resizebox{0.5\linewidth}{!}{
%     \begin{tabular}{c|ccc}
%         \toprule
%         Method  & Clotho-AQA & AudioCaps & AudioSet-SL  \\ \toprule
%         Video LLaMA&x&x&x  \\
%         MacawLLM&x&x&x  \\
%         AVLLM&x&x&x  \\
%         GroundingGPT&x&x&x \\
%         \modelname (Ours)&x&x&x \\
    
%          \bottomrule    
%         \end{tabular}}
%     \label{tab:audio_results}
% \end{table*}

\section{Limitations and Future Work}
Here, we outline some limitations that are important considerations for future work.

First, the \texttt{OCTAV} dataset consists of sounds that are non-overlapping and distinct, which simplifies the learning and classification process. However, in real-life scenarios, sound events often overlap, occur simultaneously, and can be highly ambiguous. This makes sound detection and classification far more complex. Thus, a natural extension of our work would be to incorporate sound data that reflects more in-the-wild conditions, where sounds are less controlled, overlap frequently, and can exhibit high variability in intensity and duration. Adapting the dataset to represent such real-world complexities will enhance the robustness and applicability of the model in practical applications.

Second, our proposed \texttt{OMCAT} model employs the CLIP visual encoder~\citep{radford2021learning} as the video encoder, which focuses on frame-based visual representations. While CLIP has demonstrated strong capabilities in multimodal learning, it lacks explicit modeling of temporal dynamics between video frames. Given that many real-world events are temporally dependent—especially in video sequences—using a video-based encoder that captures temporal consistency, such as those designed for action recognition~\citep{ren2024timechat}, would likely result in more accurate and nuanced representations of events. In future work, we aim to explore alternative video encoders that model temporal aspects of video more effectively, enabling better alignment between the visual and audio modalities in complex, dynamic environments. This could lead to more sophisticated models capable of handling temporal dependencies and multi-event interactions in both visual and audio data.

Third, currently the dataset consists of short-length videos ($\sim$30-40 seconds), extending the dataset to long videos would be extremely beneficial for practical applications. Longer videos would provide more comprehensive context, allowing models to better capture temporal dependencies, complex patterns, and interactions that unfold over extended periods. Moreover, long-duration videos would enable more robust testing and evaluation in real-world scenarios, where short clips often fail to represent the full dynamics of real-time events. Expanding the dataset in this way would lead to more accurate models and improve their generalizability across a broader range of applications.